\documentclass[twoside,fleqn]{article}

\usepackage{amsmath}
\usepackage{amssymb}
\usepackage{wasysym}
\usepackage[accepted]{aistats2015}

\usepackage{hyperref}
\usepackage{url}
\usepackage{graphicx}
\usepackage[scaled=0.95]{inconsolata}
\usepackage{mathrsfs}
\usepackage{natbib}

\usepackage{algpseudocode}

\let\olditemize\itemize
\renewcommand{\itemize}{
  \olditemize
  \setlength{\itemsep}{4pt}
  \setlength{\parskip}{0pt}
  \setlength{\parsep}{0pt}
}

\newcommand{\p}[2]{\ensuremath{p(#1 \,|\, #2)}}

\renewcommand{\vec}[1]{\ensuremath{\boldsymbol{#1}}}

\newcommand{\R}{\ensuremath{\mathcal{R}}}
\renewcommand{\L}{\ensuremath{\mathcal{L}}}
\newcommand{\T}{\ensuremath{\mathcal{T}}}

\newcommand{\x}{\ensuremath{\vec{x}}}
\newcommand{\y}{\ensuremath{\vec{y}}}

\begin{document}

\twocolumn[

\aistatstitle{Particle Gibbs with Ancestor Sampling for Probabilistic Programs}

%

\aistatsauthor{~~~~~~~ Jan-Willem van de Meent \And ~~~~~~~ Hongseok Yang \And ~~~~~~~~ Vikash Mansinghka \And  ~~~~~ Frank Wood}

\aistatsaddress{~~~~~~~~~ Dept of Statistics \\ ~~~~~~~~~  Columbia University \And ~~~~~~~ Dept of Computer Science \\ ~~~~~~~ University of Oxford \And ~~~~~~~~ Computer Science \& AI Lab \\ ~~~~~~~~ Mass Institute of Technology \And ~~~~~ Dept of Engineering \\ ~~~~~ University of Oxford} ]

\begin{abstract}

Particle Markov chain Monte Carlo techniques rank among current state-of-the-art  methods for probabilistic program inference.
A drawback of these techniques is that they rely on importance resampling, which results in  degenerate particle trajectories and a low effective sample size for variables sampled early in a program. 
We here develop a formalism to adapt ancestor resampling, a technique that mitigates particle degeneracy, to the probabilistic programming setting.
We present empirical results that demonstrate nontrivial performance gains.

%
\end{abstract}

\section{Introduction}


Probabilistic programming languages extend traditional languages with primitives for sampling and conditioning on random variables.
Running a program $F$ generates random variates $\x$ for some subset of program expressions, which can be thought of as a sample from a prior $\p{\x}{F}$ implicitly defined by the program. 
In a conditioned program $F[\vec{y}]$, a subset of expressions is constrained to take on observed values $\y$. 
This defines a posterior distribution $\p{\x}{F[\vec{y}]}$ on the random variates $\x$ that can be generated by the program $F[\vec{\y}]$.

Probabilistic programs are in essence procedural representations of generative models. 
These representations are often both succinct and extendable, making it easy to iterate over alternative model designs. 
Any program that always samples and constrains a fixed set of variables $\{\x,\y\}$ admits an alternate representation as a graphical model.
In general, a program $F[\y]$ may also have random variables that are only generated when certain conditions on previously sampled variates are met. 
The random variables $\x$ therefore need not have the same entries for all possible executions of $F[\y]$.
In languages that have recursion and higher-order functions (i.e. functions that act on other functions), it is straightforward to define models that can instantiate an arbitrary number of random variables, such as certain Bayesian nonparametrics, or models that are specified in terms of a generative grammar.
At the same time this greater expressivity makes it challenging to design methods for efficient posterior inference in arbitrary programs.

In this paper we show how a recently proposed technique known as particle Gibbs with ancestor sampling (PGAS) \cite[]{lindsten_nips_2012} can be adapted to inference in higher-order probabilistic programming systems \cite[]{mansinghka_arxiv_2014,wood_aistats_2014,goodman_uai_2008}.
A PGAS implementation requires combining a partial execution history, or prefix, with the remainder of a previously completed execution history, which we call a suffix. 
We develop a formalism for performing this operation in a manner that guarantees a consistent program state, correctly updates the probabilities associated with each sampled and observed random variable, and avoids unnecessary recomputation where possible. 
An empirical evaluation demonstrates that the increased statistical efficiency of PGAS can easily outweigh its greater computational cost.



\subsection{Related Work}

Current generation probabilistic programming systems fall into two broad categories. 
On the one hand, systems like Infer.NET \cite[]{minka_software_2010} and STAN \cite[]{stan_software_2014} restrict the language syntax in order to omit recursion. 
This ensures that the set of variables is bounded and has a well-defined dependency hierarchy.
On the other hand languages such as Church \cite[]{goodman_uai_2008}, Venture \cite[]{mansinghka_arxiv_2014}, Anglican \cite[]{wood_aistats_2014}, and Probabilistic C \cite[]{paige_icml_2014} do not impose such restrictions, which makes the design of general-purpose inference methods more difficult.

Arguably the simplest inference methods for probabilistic programming languages rely on Sequential Monte Carlo (SMC) \cite[]{delmoral_jrssb_2006}. 
These ``forward'' techniques sample from the prior by running multiple copies of the program, calculating importance weights using the conditioning expressions as needed. 
The only non-trivial requirement for implementing SMC variants is that there exists an efficient mechanism for ``forking'' a program state into multiple copies that may continue execution independently.

Another well-known inference technique for probabilistic programs is lightweight Metropolis-Hasting (LMH) \cite[]{wingate_aistats_2011}. 
LMH methods construct a database of sampled values at run time. 
A change to the sampled values is proposed, and the program is rerun in its entirety, substituting previously sampled values where possible.
The newly constructed database of random variables is then either accepted or rejected.
LMH is straightforward to implement, but costly, since the program must be re-run in its entirety to evaluate the acceptance ratio.
A more computationally efficient strategy is offered by Venture \cite[]{mansinghka_arxiv_2014}, which represents the execution history of a program as a graph, in which each evaluated expression is a node. 
A graph walk can then determine the subset of expressions affected by a proposed change, allowing partial re-execution of the program conditioned on all unaffected nodes.

SMC and MH based algorithms each have trade-offs. 
Techniques that derive from SMC can be run very efficiently, but suffer from particle degeneracy, resulting in a deteriorating quality of posterior estimates calculated from values sampled early in the program.  
In MH methods subsequent samples are typically correlated, and many updates may be needed to obtain an independent sample. 
As we will discuss in Section~\ref{sec:pmcmc}, PGAS can be thought of as a hybrid technique, in the sense that SMC is used to generate independent updates to the previous sample, which mitigates the degeneracy issues associated with SMC whilst increasing mixing rates relative to MH.

\section{Probabilistic Functional Programs}


\subsection{Language Syntax}

For the purposes of exposition we will consider a simple Lisp dialect, extended with primitives for sampling and observing random variables
\begin{verbatim}
  e ::= c | s | (e &e) | (lambda (&s) e) 
        | (if e e e) | (quote e) 
  v ::= bool | int | float | string | primitive 
        | compound | (&v) | stochastic 
\end{verbatim}

An expression {\tt e} is either a constant literal {\tt c}, a symbol {\tt s}, an application {\tt (e \&e)} with operator {\tt e} and zero or more arguments {\tt \&e}, a function literal {\tt (lambda (\&s) e)} with argument list {\tt (\&s)}, an if-statement {\tt (if e e e)}, or a quoted expression {\tt (quote e)}. 
Each expression {\tt e} evaluates to a value {\tt v} upon execution. 
In addition to the self-explanatory {\tt bool}, {\tt int}, {\tt float}, and {\tt string} types, values can be {\tt primitive} procedures (i.e. language built-ins such {\tt +}, {\tt -}, etc.), {\tt compound} procedures (i.e., closures), and lists of zero or more values {\tt (\&v)}. 

The {\tt stochastic} type represents stochastic processes, whose samples must either be i.i.d.~or exchangeable.
A stochastic process {\tt sp} supports two operations
\begin{verbatim}
  (sample sp) -> v
  (observe sp v) -> v
\end{verbatim}
The {\tt sample} primitive draws a value from {\tt sp}, whereas the {\tt observe} primitive conditions execution on a sample {\tt v} that is returned as passed.
Both operations associate a value {\tt v} to {\tt sp} as a side-effect, which changes the probability of the execution state in the inference procedure. 
For exchangeable processes such as {\tt (crp 1.0)} this also affects the probability of future samples.

Following the convention employed by Venture and Anglican, we define programs as sequences of three types of top-level statements
\begin{verbatim}
  F ::= t F | END
  t ::= [assume s e] | [observe e v] | [predict e]
\end{verbatim}
Variables in the global environment are defined using the {\tt assume} directive.
The {\tt observe} directive conditions execution in the same way as its non-toplevel equivalent. 
The {\tt predict} directive returns the value of the expression {\tt e} as inference output. 

\subsection{Importance Sampling Semantics}

In many probabilistic languages the {\tt (observe e v)} form is semantically equivalent to imposing a rejection-sampling criterion.
For example, a program subject to {\tt (observe (> a 0) true)} can be interpreted as a rejection sampler that repeatedly runs the program and only returns {\tt predict} values when {\tt (> a 0)} holds.

The semantics of {\tt observe} that we have defined here imply an interpretation of a probabilistic program as an importance sampler, where {\tt sample} draws from the prior and {\tt observe} assigns an importance weight.
Instead of constraining the value of an arbitrary expression {\tt e}, {\tt observe} conditions on the value of {\tt (sample e)}.
This restricted form of conditioning guarantees that we can calculate the likelihood $p(${\tt v}$\,|\,${\tt (sample e)}$)$ for every {\tt observe}, as long as we implement a density function for all possible {\tt stochastic} values in the language.

More formally, we use the notation $F[\y]$ to refer to a program conditioned on values $\y$ via a sequence of top-level {\tt [observe e v]} statements. 
Execution of $F[\y]$ will require the evaluation of a number of {\tt (sample e)} expressions, whose values we will denote with $\x$.
We now informally define $F[\y,\x]$ as the program in which all {\tt (sample e)} expressions are replaced by conditioned equivalents {\tt (observe e v)}, resulting in a fully deterministic execution.
Similarly we can define $F[\x]$ as the program obtained from $F[\y,\x]$ by replacing {\tt [observe e v]} statements with 
unconditioned forms {\tt [assume s (sample e)]} with a unique symbol {\tt s} in each statement.

Whereas top-level {\tt observe} statements are fixed in number and order, $F$ may not evaluate the same combination of {\tt sample} calls in every execution.
To provide a more precise definition of $F[\x]$, we associate a unique address $\alpha$ with each {\tt sample} call that can occur in the execution of $F$.
We here use a scheme in which the run-time address of each evaluation is a concatenation $\alpha'\!\!::\!\!(t,p)$ of the address $\alpha'$ of the parent evaluation and a tuple $(t,p)$ in which $t$ identifies the expression type and $p$ is the index of the sub-expression within the form. 
This particular scheme labels every evaluation, not just the sample calls, allowing us to formally represent $F:\mathcal{A} \to \mathcal{E}$ as a mapping from addresses $\mathcal{A}$ to program expressions {$\mathcal{E}$.
We represent $\x:\mathcal{S} \to \mathcal{V}$ as a mapping from a subset of addresses $\mathcal{S} \subset \mathcal{A}$ associated with {\tt sample} calls to values $\mathcal{V}$. 
Similarly, $\y:\mathcal{O} \to \mathcal{V}$ is a mapping from addresses $\mathcal{O} \subset \mathcal{A}$ associated with {\tt observe} statements to values.
With these definitions in place, a conditioned program $F[\x]: \mathcal{A} \to \mathcal{E}$ simply replaces $F(\alpha) =\:${\tt (sample e)} with
\[
	F[\x](\alpha) =\:${\tt (observe e }$\x(\alpha)${\tt)}$, 
	\qquad 
	\forall\alpha \in \mathcal{S}.
\]
The importance sampling interpretation of a program $F[\y] \leadsto W, \x$ (read as $F[\x]$ yields $W,\x$) is defined in terms of random variables $\x$ and a weight $W$
\begin{align*}
  F[\y] &\leadsto W,\x
  &
  W &= \p{\y}{F[\x]}
  ~.
\end{align*}
By definition, the generated samples $\x$ are drawn from the prior $\p{\x}{F}$. 
The weight $W=\p{\y}{F[\x]}$ is the joint probability of all top-level {\tt observe} statements in $F[\y]$.
Repeated execution of $F[\y]$ yields a weighted sample set $\{W^l,\x^l\}$ that may be used to approximate the posterior as 
\begin{align*}
  \p{\x}{F[\y]}
  &
  \simeq
  p^L(\x \,|\, F[\y])
  =
  \sum_l 
  \frac{W ^l}
       {\textstyle \sum_k W^k}
  \delta_{\x^l}
  .
\end{align*}
More generally the importance weight is defined as the joint probability of all {\tt observe} calls (top-level and transformed) in the program
\begin{align*}
  F[\y,\x] &\leadsto W 
  &
  W &= \p{\y,\x}{F}
  ,
  \\
  F[\x] &\leadsto W, \y
  &
  W &= \p{\x}{F[\y]}
  ,
  \\
  F &\leadsto W,\y,\x
  &
  W &= 1
  .
\end{align*}

\section{Particle MCMC methods}
\label{sec:pmcmc}


\subsection{Sequential Monte Carlo}

SMC methods are importance sampling techniques that target a posterior $\p{\vec{x}}{\vec{y}}$ on as space $\mathcal{X}$ by performing importance sampling on unnormalized densities $\{\gamma_n(\vec{x}_n)\}_{n=1}^N$ defined on spaces of expanding dimensionality $\{\mathcal{X}_n\}_{n=1}^N$, where each $\mathcal{X}_n \subseteq \mathcal{X}_{n+1}$ and $\mathcal{X}_N = \mathcal{X}$.
This results in a series of intermediate particle sets $\{w_n^l,\vec{x}_n^l\}_{n=1}^N$, which we refer to as generations. 
Each generation is sampled via two steps,
\begin{align*}
	a_n^l 
	&\sim  
	R(a_n \,|\, w_n)
	,
	&
	\x_n^l 
	&\sim 
	\rho_n(\x_n \,|\, \x_{n-1}^{a_n^l})
	.
\end{align*}
Here $R(a \,|\, w)$ is a resampling procedure that returns index $a=l$ with probability $w^l / \sum_{l'} w^{l'}$ and $\rho_n$ is a transition kernel. 
The samples $\x_n^l$ are assigned weights
\begin{equation*}
	w^l_n = \frac{\gamma_n(\vec{x}_n^l)}
	             {\gamma_{n-1}(\vec{x}_{n-1}^{a_n^l}) \rho_n(\vec{x}^l_n \,|\, \vec{x}_{n-1}^{a_n^l})}
	~.             
\end{equation*}
In the context of probabilistic programs, we can define a series of partial programs $F_n[\y_n]$ that truncate at each top-level {\tt [observe e v]} statement. 
We can then sequentially sample $F_n[\y_n,\x_{n-1}] \leadsto W_n,\x_n$ by partially conditioning on $\x_{n-1}$ at each generation. 
Since $\x_n$ is a sample from the prior, this results in an importance weight \cite[]{wood_aistats_2014}
\begin{align*}
	w_n^l
	&= \frac{\p{\y_n}{F_n[\x_n^l]}}
	        {\p{\y_{n-1}}{F_{n-1}[\x_{n-1}^{a_n^l}]}}
	.
\end{align*}
Note that this is simply the likelihood of the $n$-th top-level {\tt observe}.
In practice we continue execution relative to $F_{n-1}[\y_{n-1},\x_{n-1}]$ to avoid rerunning $F_n[\y_n,\x_{n-1}]$ in its entirety. 
The means that the inference backend must include a routine for forking multiple independent executions from a single state.

\subsection{Iterative Conditional SMC}

An advantage of SMC methods is that they provide a generic strategy for joint proposals in high dimensional spaces. 
An importance sampling scheme where $F[\y] \leadsto W,\x$ draws from the prior has a vanishingly small probability of generating a high-weight sample.
By sampling the smallest possible set of variables $\vec{x}^l_n$ at each generation and selecting ancestors $a^l_{n+1}$ according to the likelihood of the next observed data point, we ensure that $\x^l_{n+1}$ is sampled conditioned on high-weight values of $\x_{n}$ from the previous generation.
At the same time this strategy has a drawback: each time the particle set is resampled, the number of unique values at previous generations decreases, typically resulting in coalescence to a single common ancestor in $O(L \log L)$ generations \cite[]{jacob_sc_2013}.

In many applications it is not practically possible (due to memory requirements) to set $L$ to a value large enough to guarantee a sufficient number of independent samples at all generations.
In such cases, particle variants of MCMC techniques \cite[]{andrieu_jrssb_2010} can be used to combine samples from multiple SMC sweeps.
An iterated conditional SMC (ICSMC) sampler repeatedly selects a retained particle $k$ with probability $w_N^k / \sum_{k'} w_N^{k'}$, and then performs a conditional SMC (CSMC) sweep, where the resampling step is conditioned on the inclusion of the retained particle at each generation.
Formally, this procedure is a partially collapsed Gibbs sampler that targets a density $\phi(\vec{x}, a, k)$ on an extended space
\begin{align*}
	&
	\phi(\x^{1:L}_{1:N}, a^{1:L}_{2:N}, k) 
	=
	\frac{w_N^k}{\textstyle \sum_{k'} w_N^{k'}}
	\prod_{l=1}^L 
	\p{\x_1^l}{F_1}
	\\
	&
	\qquad
	\prod_{n=2}^N
	\prod_{l=1}^L
	\frac{w_{n-1}^{a_n^l}}{\textstyle \sum_{l'} w_{n-1}^{l'}}
	\p{\vec{x}_n^l}{F_n[\vec{x}_{n-1}^{a^l_n}]}
	~.
\end{align*}
We use the shorthand $\vec{x}^k = \x^{b_1:b_N}_{1:N}$ and $a^k = a^{b_2:b_N}_{2:N}$ to refer to the sampled values and ancestor indices of the retained particle, whose index $b_n$ at each generation can be recursively defined via $b_N=k$ and $b_{n-1}=a_n^{b_n}$. 
The notation $\vec{x}^{-k}$ and $a^{-k}$ refers to the complements where the retained particle is excluded.
An ICSMC sampler iterates between two updates
\begin{itemize}
\item[1.] $\{\vec{x}^{*,-k}, a^{*,-k}\} \sim \phi(\vec{x}^{-k}, a^{-k} \,|\, \vec{x}^k, a^k, k)$
\item[2.] $k^{*} \sim \phi(k \,|\, \vec{x}^{*,-k}, a^{*,-k}, \vec{x}^k, a^k)$
\end{itemize}
If we interpret $\x^{-k}$, $a$ and $k$ as auxilliary variables, then the marginal on $\x_N^k$ leaves the density $\gamma_N(\x_N)$ invariant \cite[]{andrieu_jrssb_2010}.

The advantage of ICSMC samplers is that the target space is iteratively explored over subsequent CSMC sweeps.
A disadvantage is that consecutive sweeps often yield partially degenerate particles, since many newly generated particles will coalesce to the retained particle with high probability.
For this reason ICSMC samplers mix poorly when the number of particles is not large enough to generate at least two completely independent lineages in a single CSMC sweep.

\subsection{Particle Gibbs with Ancestor Sampling}

PGAS is a technique that augments the CSMC sweep with a resampling procedure for the index $a_n^{b_n}$ of the retained particle \cite[]{lindsten_nips_2012}. 
At a high level, this sampling scheme performs two updates
\begin{itemize}
\item[1.] $\{\vec{x}^{*,-k}, a^{*}\} \sim \phi(\vec{x}^{-k}, a \,|\, \vec{x}^k, k)$
\item[2.] $k^* \sim \phi(k \,|\, \vec{x}^{*,-k}, a^{*,-k}, \vec{x}^k, a^k)$
\end{itemize}
Here update 1 differs from the normal CSMC update in that it samples a complete set of ancestor indices $a^*$, not the complement to the retained indices $a^{*,-k}$. 
At each generation the ancestor $a_n^{b_n}$ of the retained particle is resampled according to a weight
$$
	w^{l}_{n-1|N} 
	= 
	w_{n-1}^l 
	\p{\y_N,\x^k_N[\x^{l}_n]}{F_N[\y_n,\x^{l}_n]}.
$$
Here $\x_N^k[\x_n^{l}]$ denotes the substitution of $\x_n^{l}$ into $\x_N^k$, which is defined as a mapping $\x_N^k[\x_n^{l}] : \mathcal{A}_N^k \cup \mathcal{A}_n^l \to \mathcal{V}$ where entries in $\x_n^{l}:\mathcal{A}_n^l \to \mathcal{V}$ augment or replace entries in $\x_N^k :\mathcal{A}_N^k \to \mathcal{V}$.

Intuitively, ancestor resampling can be thought of as proposing new program executions by complementing random variables $\x_n^l$ of a partial execution, or prefix, with retained values from $\x_N^k$ to specify the future of the execution, or suffix.
This step is performed at each generation, allowing the retained lineage to potentially be resampled many times in the course of one sweep.

Incorporation of the ancestor resampling step results in the following updates at each $n = 2, \ldots, N$
\begin{itemize}
\item[1a.] Update the retained particle
\begin{align*}
	a_n^{*,b_n} 
	&\sim 
	R(a_n \,|\, w^{*}_{n-1|N})
	\\
	\vec{x}^{*,b_n}_n
	&
	\leftarrow
	\vec{x}^{b_n}_n
\end{align*}
\item[1b.] Update the particles for $l \in \{1,...,L\} \setminus b_n$
\begin{align*}
 	a_n^{*,l}
 	&\sim 
 	R(a_n \,|\, w^{*}_{n-1})
 	\\
 	\vec{x}_n^{*,l}
 	&\sim 
 	p(\vec{x}_n \,|\, F_n[\vec{x}^{*,{a^{*,l}_n}}_{n-1}])
\end{align*}
\end{itemize}

\section{Rescoring Probabilistic Programs}


PGAS for probabilistic programs requires calculation of $\p{\y_N,\x^{k}_{N}[\x_n^l]}{F[\y_n,\vec{x}^l_n]}$.
This probability is, by definition, the importance weight of $F_N[\y_N,\x^k_N[\x^{l}_n]]$ and can therefore in principle be obtained by executing this re-conditioned form.
The main drawback of this naive approach is that it requires $LN$ evaluations of the program in its entirety.
This results in an $O(LN^2)$ computational cost, which quickly becomes prohibitively expensive as the number of generations $N$ increases.
A second complicating factor is that naively rewriting part of the execution history of the program may not yield a set of random values $\x_N^k[\vec{x}^l_n]$ that could be generated by running $F_N[\y_N]$.

We here develop a formalism that allows regeneration of a self-consistent program execution starting from a partial execution with random variables $\x_n^l$, assuming all future random samples are inherited from retained values $\x_N^k$.
To do so we introduce the notion of a trace, a data structure that annotates each evaluation with information necessary to re-execute the expression relative to a new program state.
Given a trace for the suffix, i.e.~the remaining top-level statements in a program, it becomes possible to re-execute conditioned on future random values, in a manner that avoid unnecessary recomputation where possible.


\subsection{Intuition}

In higher order languages with recursion and memoization, regeneration is complicated by two factors:

\begin{itemize}
\item[1.] The program $F_N[\y_N, \x^k_N[\x^{l}_n]]$ may be underconditioned, in the sense that $\x^k_N[\x^{l}_n]$ does not contain values for some {\tt sample} calls that can be evaluated in its execution.
It can also be overconditioned, when $\x^k_N[\x^{l}_n]$ contains values for {\tt sample} calls that will never be evaluated. 

\item[2.] The expression for the {\tt stochastic} argument to each {\tt observe} in 
$F_N[\y_N, \x^k_N[\x^{l}_n]]$ may need to be re-evaluated if it in some way depends on global variables defined in $F_n[\y_n,\vec{x}^l_n]$. 
Because each variable may in turn reference other variables, we must be able to reconstruct the program environment recursively in order to rescore each {\tt observe}.
\end{itemize}
The first non-triviality arises from the existence of {\tt if} expressions. 
As an example, consider the program
{\small
\begin{verbatim}
  0: [assume random? (sample (flip-dist 0.5))]
  1: [assume mu (if random?
                 (sample (normal-dist 0.0 1.0))
                 0.0)]
  2: [observe (normal-dist mu 1.0) 0.1]
\end{verbatim}}

\normalsize
The {\tt sample} expression in line {\tt 1} is only evaluated when the {\tt sample} expression in line {\tt 0} evaluates to {\tt true}.
More generally, any program that contains {\tt (sample e)} inside an {\tt if} expression will not be guaranteed to instantiate the same random variables in cases where the predicate of the {\tt if} expression itself depends on previously sampled values. 
In programming languages that lack recursion we have the option of evaluating both branches and including or excluding the associated probabilities conditioned on the predicate value.
This is essentially the strategy that is employed to handle {\tt if} expressions in Infer.NET and BUGS variants.
In languages that do permit recursion, this is in general not possible. 
For example, the following program would require evaluation of an infinite number of branches:

{\small
\begin{verbatim}
  0: [assume geom (lambda (p)
                    (if (sample (flip-dist p))
                      1
                      (+ 1 (geom p))))]
  1: [observe (poisson-dist (geom 0.5)) 3]
\end{verbatim}}

\normalsize
In other words, we cannot in general pre-evaluate the values associated with both branches in the suffix. 
When a predicate in the suffix no longer takes on the same value, we have a choice of either rejecting the regenerated suffix outright, or updating it using a regeneration procedure that evaluates the newly chosen branch and removes reference to any values sampled in the invalidated branch.
We here consider the former strict form of regeneration, which guarantees that the regenerated suffix references precisely the same set of sample values as before.

{\normalsize A second aspect that complicates rescoring is the existence of memoized procedures. As an example, consider the following infinite mixture model}
{\small
\begin{verbatim}
  0: [assume class-prior (crp 1.0)]
  1: [assume class 
       (mem (lambda (n) 
         (sample class-prior)))]
  2: [assume class-dist 
       (mem (lambda (k) 
         (normal-dist 
           (sample (normal-dist 0.0 1.0)) 
           1.0)))]
  3: [observe (class-dist (class 0)) 2.1]
  4: [observe (class-dist (class 1)) 0.6]
  ...
  N+3: [observe (class-dist (class N)) 1.2]

\end{verbatim}}
\normalsize
Here each {\tt observe} makes a call to {\tt class}, which samples an integer class label {\tt k} from a Chinese restaurant Process (CRP). 
The call to {\tt class-dist} either retrieves an existing {\tt stochastic} value, or generates one when a new value {\tt k} is encountered.
This type of memoization pattern allows us to delay sampling of the parameters until they are in fact required in the evaluation of a top-level {\tt observe}, and makes it straightforward to define open world models with unbounded numbers of parameters.
At the same time it complicates analysis when performing rescoring.
Memoized procedure calls are semantically equivalent to lazily defined variables. 
Programs that rely on memoization can therefore essentially define variables in a non-deterministic order. 
A regeneration operation must therefore dynamically determine the set of variables that need to be re-evaluated at run time.



\subsection{Traced Evaluation}

The operations that need to happen during a rescoring step are (1) the regeneration of a consistent set of global environment variables, which includes any bindings in the prefix, augmented with any bindings defined in the suffix (some of which may require re-evaluation as a result of changes to bindings in the prefix), (2) a verification that the flow control path in the suffix is consistent with the environment bindings in the prefix, and (3) the recomputation of the probabilities of any sampled and observed values whose density values depend on bindings in the prefix.

In order to make it possible to perform the above operations, we begin by introducing a set of annotations for each value {\tt v} that is returned upon evaluation of an expression {\tt e}. 
In practical terms, each value in the language is \emph{boxed} into a data structure which we call a trace.
We represent a trace $\tau$ as tuple $({\tt v},\epsilon,l,\rho,\omega,\sigma,\phi)$.  
{\tt v} is the value of the expression.
$\epsilon$ is a partially evaluated expression, whose sub-expressions are themselves represented as traces.
$l$ is the accumulated log-weight of {\tt observes} evaluated within the expression.
$\rho$ is a mapping $\{{\tt s} \to \tau\}$ from symbols to traces, containing the subset of the global environment variables that were referenced in the evaluation of {\tt e}. 
$\omega$ is a mapping $\{\alpha \mapsto (\tau,{\tt v},l)\}$. It contains an entry at the address $\alpha$ of each {\tt observe} that was evaluated in {\tt e} and its sub-expressions.
This entry is represented as a tuple $(\tau,{\tt v},l)$ containing a trace of the first 
argument to the {\tt observe} (which must be of the {\tt stochastic} type),
the observed value {\tt v}, and the associated log-weight $l$.
Similarly $\sigma$ is a mapping $\{\alpha \mapsto (\tau,{\tt v})\}$ that contains an entry for each evaluated {\tt sample} expression (which omits the associated log-weight). 
The last component $\phi$ is again a mapping $\{\alpha \mapsto \tau\}$ that records all traces that appear as conditions in {\tt if} expressions and thereby influence the control flow of program execution.

We now describe the semantics of the traced evaluation $({\tt e},\alpha,R,\Lambda) \Downarrow \tau$. 
The evaluation operator $\Downarrow$ returns the trace $\tau$ of an expression ${\tt e}$ at address $\alpha$, relative to a global environment $R$ (i.e.~variables defined via {\tt assume} statements) and local bindings $\Lambda$ (i.e.~variables bound in compound procedure calls), resulting in a trace $\tau = ({\tt v},\epsilon,l,\rho,\omega,\sigma,\phi)$.
We assume the implementation provides a standard evaluation function for primitive procedures 
$\mathit{eval}(\mathtt{prim}~\mathtt{v}_1 \ldots \mathtt{v_n}) = \mathtt{v}$.
We reiterate that the notation $\alpha{::}(t,p)$ denotes an evaluation address composed of a parent address $\alpha$, a type identifier $t$ and a sub-expression index $p$, where $t$ is one of {\tt i} for {\tt if}, {\tt l} for {\tt lambda}, {\tt q} for {\tt quote}, {\tt a} for applications, and {\tt b} when evaluating compound procedure bodies.

Constants {\tt c} evaluate to:
\begin{align*}
  ({\tt c}, \alpha, R, \Lambda) 
  &\Downarrow
  ({\tt c}, {\tt c},0.0,\{\},\{\},\{\},\{\}) 
  ~.
\end{align*}
We call this type of trace ``transparent'', since it contains no references to other traces that may take on different values or probabilities in another execution.

Symbol lookups {\tt s} in the global environment return the value of {\tt s} stored in $R$:
\begin{align*}
  &({\tt s}, \alpha, R, \Lambda) 
  \Downarrow
  ({\tt v}, {\tt s},0.0,\{{\tt s} \mapsto \tau\},\{\},\{\},\{\})
  \\
  &
  \quad 
  {\rm if}~ R({\tt s}) = \tau ~\text{and}~ \tau = ({\tt v},\_,\_,\_,\_,\_,\_)  
  ~.
\end{align*}
Lookups from the local environment are inlined:
\begin{align*}
  &({\tt s}, \alpha, R, \Lambda) 
  \Downarrow
  ({\tt v},\epsilon,0.0,\rho,\omega,\sigma,\phi)
  \\
  &\quad
  {\rm if}~ \Lambda({\tt s}) = \tau ~\text{and}~ \tau = ({\tt v},\epsilon,l,\rho,\omega,\sigma,\phi)
  ~.  
\end{align*}
Calls to {\tt sample} inherit annotations from the trace $\tau$ passed as an argument, and add an entry in $\sigma$:
\begin{align*}
  &
  (({\tt sample} ~{\tt e}), \alpha, R, \Lambda)
  \,{\Downarrow}\, ({\tt v}, {\tt v}, l, \rho, \omega, \sigma[\alpha \,{\mapsto}\, (\tau,{\tt v})],\phi)
  \\
  &\quad
  \begin{array}{@{}l@{}}
  \mbox{if $({\tt e}, \alpha {::} ({\tt s}, 0), R, \Lambda) \Downarrow \tau = ({\tt v}_1, \_, l, \rho, \omega, \sigma, \phi)$ and}
  \\
  \ \mbox{{\tt v} is drawn from the {\tt stochastic} process$~{\tt v}_1$}~.
  \end{array}
\end{align*}
Calls to {\tt observe} inherit from $\tau$ and add an entry in $\omega$:
\begin{align*}
  &
  (({\tt observe} ~{\tt e}_1~{\tt v}_2), \alpha, R, \Lambda)
  \\
  &\Downarrow ({\tt v}_2, {\tt v}_2, l_1 + l_{12},\rho,\omega[\alpha \mapsto (\tau,{\tt v}_2,l_{12})],\sigma,\phi)
  \\
  &
  \begin{array}{@{}l@{}}
  \quad{\rm if}~({\tt e}_1, \alpha {::} ({\tt o}, 0), R, \Lambda) \Downarrow \tau = ({\tt v}_1, \_, l_1, \rho, \omega, \sigma, \phi))
  \\
  \quad\quad \mbox{and}~l_{12} = \L({\tt v}_1,{\tt v}_2)~.
  \end{array}
\end{align*}
Here 
$\L({\tt v}_1,{\tt v}_2)$ is used to denote the log-density of value ${\tt v}_2$ relative to ${\tt v}_1$ (which must be of type {\tt stochastic}). 

Primitive procedure applications $({\tt primop}~{\tt e}_1~{\tt e}_2)$ evaluate to 
$\mathit{eval}({\tt prim}~{\tt v}_1~{\tt v}_2)$ where ${\tt v}_i$ is the result of evaluating ${\tt e}_i$:
\begin{align*}
  &
  (({\tt prim}~{\tt e}_1~{\tt e}_2),\alpha,R,\Lambda)
  \\
  &\Downarrow
  (\mathit{eval}({\tt prim}~{\tt v}_1~{\tt v}_2),({\tt prim}~\tau_1~\tau_2),l,\rho,\omega,\sigma,\phi)~,
\end{align*}
if $({\tt e}_i, \alpha{::}({\tt p}, i), R, \Lambda) \Downarrow \tau_i$, and
$\rho$, $\omega$, $\sigma$, and $\phi$ are obtained by merging the corresponding components of the $\tau_i$,
and the log-density $l$ is the sum of the $l_i$.

Application of a single-argument compound procedure (i.e. closure) ${\tt e}_1$
leads to the evaluation of the body {\tt e} of the procedure
relative to the environments $R_1,\Lambda_1$ in which the compound procedure was defined:
\begin{align*}
  &
  (({\tt e}_1~{\tt e}_2),\alpha,R,\Lambda)
  \Downarrow
  ({\tt v},(\tau_1~\tau_2),l,\rho,\omega,\sigma,\phi)
\end{align*}
if 
$$
\begin{array}{@{}l@{}}
({\tt e}_1,\alpha{::}({\tt a},0),R,\Lambda) \Downarrow \tau_1,
\\
\tau_1 = ({\tt (lambda ~(s) ~e)}, R_1, \Lambda_1),\_,\_,\_,\_,\_,\_),
\\
({\tt e}_2,\alpha{::}({\tt a},1),R,\Lambda) \Downarrow \tau_2 = ({\tt v}_2,\_,\_,\_,\_,\_,\_),
\\
({\tt e}, \alpha{::}({\tt b}, 0), R_1, \Lambda_1[{\tt s} \mapsto \tau_2]) \Downarrow \tau_3 = ({\tt v}, \_, \_, \_,\_,\_,\_),
\end{array}
$$
and $l,\rho,\omega,\sigma,\phi$ are obtained by combining the corresponding components of the $\tau_i$.
The case with multiple arguments is defined similarly.

Quote expressions $({\tt quote}~\tau)$ simply return $\tau$:
$$
  (({\tt quote}~{\tt e}),\alpha,R,\Lambda) \Downarrow \tau
  \quad
  \mbox{if}~({\tt e}, \alpha::({\tt q},0), R, \Lambda) \Downarrow \tau~.
$$

Finally, an {\tt if} expression $({\tt if}~{\tt e}~{\tt e}_1~{\tt e}_2)$ returns either 
the result of evaluating ${\tt e}_1$ or ${\tt e}_2$. We show only the case where
the {\tt true} branch is taken:
\begin{align*}
  (({\tt if}~{\tt e}~{\tt e}_1~{\tt e}_2),\alpha,R,\Lambda)
  \Downarrow
  ({\tt v},\epsilon,l,\rho,\omega,\sigma,\phi[\alpha \mapsto \tau])
\end{align*}
if 
$$
  \begin{array}[t]{@{}l@{}}
  ({\tt e},\alpha{::}({\tt i},0), R, \Lambda) \Downarrow \tau = ({\tt true},\_,\_,\_,\_,\_,\_)~,
  \\
  ({\tt e}_1, \alpha{::}({\tt i}, 1), R, \Lambda) \Downarrow \tau_1 = ({\tt v},\epsilon,\_,\_,\_,\_,\_)~,
  \end{array}
$$
and $l,\rho,\omega,\sigma,\phi$ are obtained by combining the corresponding components of $\tau$ and $\tau_1$.
The other case is that the value of $\tau$ is {\tt false}, and has
the semantics similar to the one above.



\subsection{Regeneration and Rescoring}

We now define an operation $\R(\tau, R) = \tau'$ that regenerates a traced value relative to an environment $R$. 
This operation performs the following steps: 
\begin{itemize}
\item[1.] Re-evaluate predicates: Compare $\tau=\phi(\alpha)$ to $\tau'=\R(\tau, R)$ for all $\alpha$. 
Abort if $\tau'$ and $\tau$ have different values {\tt v}. 
Otherwise update $\phi[\alpha \mapsto \tau']$.

\item[2.] Re-score {\tt observe} expressions and statements: 
Let $(\tau,{\tt v},l) = \omega(\alpha)$, and $\tau' = \R(\tau, R)$. If $\tau'$ and $\tau$
have different values ${\tt v}'_0$ and ${\tt v}_0$, recalculate $l' = \L({\tt v}'_0,{\tt v})$ and update 
$\omega[\alpha \to (\tau',{\tt v},l')]$. Otherwise, update $\omega[\alpha \to (\tau',{\tt v},l)]$.

\item[3.] Re-score samples: Let $(\tau,{\tt v}) = \sigma(\alpha)$, and $\tau' = \R(\tau, R)$. 
Calculate $l' = \L(\tau',{\tt v})$ and update $\omega[\alpha \mapsto (\tau',{\tt v},l')]$.

\item[4.] Regenerate the environment bindings: For all symbols {\tt s} that do not exist in $R$, let $\tau'=\R(\rho({\tt s}),R)$ and update $R[{\tt s} \mapsto \tau']$ and $\rho[{\tt s} \mapsto \tau']$. For existing symbols update $\rho[{\tt s} \mapsto R({\tt s})]$. We for convenience assume that $R$ is updated in place, though this may be avoided by having $\R$ return a tuple $(R',\tau')$.

\item[5.] If any bindings were changed with new values in step 4, regenerate all sub-expressions $\tau_i$ in $\epsilon$ to reconstruct $\epsilon'$.

\item[6.] If $\epsilon'$ was reconstructed in step 5, evaluate $\epsilon'$ and update {\tt v} to the result of this evaluation.
\end{itemize}

Rescoring an individual trace may be performed as part of the regeneration sweep by calculating a difference in log-density $\Delta l$.
This $\Delta l$ is the sum of all terms $l' - l$ in step 2 and all terms $l'$ in step 3, and any $\Delta l'$ values return from recursive calls to $\R$.

We have omitted a few technical details in this high-level description.
The first is that we build a map $\mathcal{C} = \{\tau \to \tau', \ldots\}$ on a call to $\R$, 
which is passed as an additional argument in recursive calls to $\R$, effectively memoizing the computation 
relative to a given initial environment $R$.
This reduces the computation on recursive calls, which potentially expand the same traces $\tau$ many times as sub-expressions of $\epsilon$.

\subsection{Ancestor Resampling}

Given an implementation of a traced evaluator and a regenerating/rescoring procedure \R, an implementation for PGAS in probabilistic programs becomes straightforward.
We represent the programs $F_n[\y_n,\x_n^l]$ evaluated up to the first $n$ top-level statements
as pairs $(R_n^l,\tau^l_n)$.
We now construct a concatenated suffix $\T_n$ by recursively re-evaluating $\T_n = ({\tt cons}~\tau^{b_n}_n~\T_{n+1})$.
For $n={1,\ldots,N}$, we then define $\p{\y_N,\x^k_N[\x_n^{*,l}]}{F_N[\y_n,\x_n^{l}]}$ as the log-weight of the rescored trace $\R(\T^l_n,R^{l}_{n-1})$.
Note here that by construction, we may extract $\T_{n+1}$ from $\T_{n}$ without additional computation.

\section{Experiments}


\begin{figure}[!t]
\includegraphics[width=3.3in]{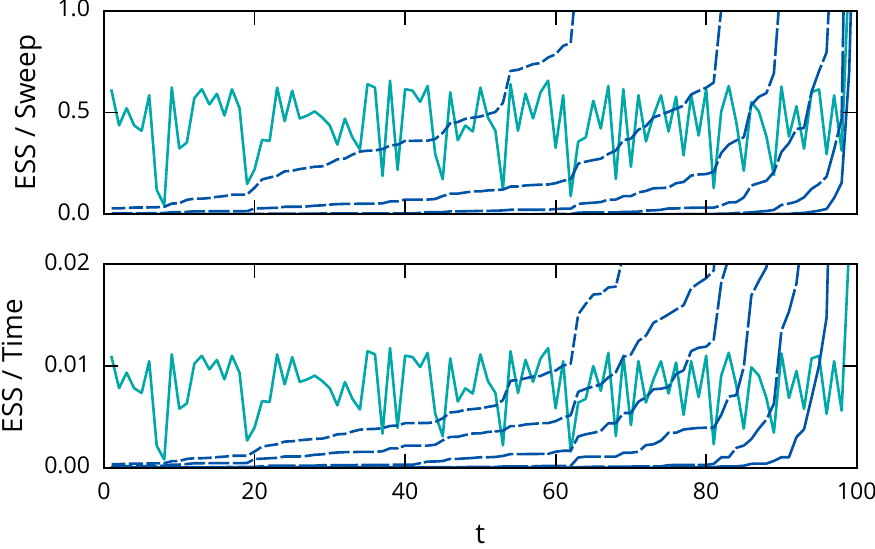}
\caption{\label{fig:lgm-ess} Effective sample size as a function of $t$, normalized by the number of PMCMC sweeps (top) and wall time in seconds (bottom). 
PGAS with 10 particles is shown in cyan. ICSMC results are shown in blue, with shorter dashes representing increasing particle counts 10, 20, 50, 100, 200, and 500.
All lines show the median ESS over 25 independent restarts.}
\end{figure}

To evaluate the mixing properties of PGAS relative to ICSMC, we consider a linear dynamical system (i.e. a Kalman smoothing problem) with a 2-dimensional latent space and a D-dimensional observational space,
\begin{align*}
    \vec{z}_t 
    &\sim 
    {\rm Norm}(\vec{A} \cdot \vec{z}_{t-1}, \vec{Q})
    ,
    &
    \y_t 
    &\sim
    {\rm Norm}(\vec{C} \cdot \vec{z}_{t}, \vec{R})
    .
\end{align*}
We impose additional structure by assuming that the transition matrix $\vec{A}$ is a simple rotation with angular velocity $\omega$, whereas the transition covariance $\vec{Q}$ is a diagonal matrix with constant coefficient $q$,
\begin{align*}
    \vec{A} 
    &= 
    \left[
        \begin{array}{cc}
            \cos \omega 
            & 
            -\sin \omega
            \\
            \sin \omega
            &
            \cos \omega
        \end{array}
    \right]
    ,
    &
    \vec{Q}
    &=
    q
    \vec{I}_2 
    .
\end{align*}
We simulate data with $D=36$ dimensions, $T=100$ time points, $\omega = 4 \pi / T$, $q=0.1$, $\alpha=0.1$, and $r=0.01$.
We now consider an inference setting where $\vec{C}$ and $\vec{R}$ are assumed known and estimate the state trajectory $\vec{z}_{1:T}$, as well as the parameters of the transition model $\omega$ and $q$, which are given mildly informative priors, $\omega \sim {\rm Gamma}(10,2.5)$ and $q \sim {\rm Gamma}(10,100)$.

While this is a toy problem where an expectation maximization (EM) algorithm could likely be derived, it is illustrative of the manner in which probabilistic programs can extend models by imposing additional structure, in this case the dependency of $\vec{A}$ on $\omega$.
This modified Kalman smoothing problem can be described in a small number of program lines

{\small
\begin{verbatim}
  [assume C [...]] ; assumed known
  [assume R [...]] ; assumed known
  [assume omega (* (sample (gamma-dist 10. 2.5))
                   (/ pi T))]
  [assume A [[(cos omega) (* -1 (sin omega))]
             [(sin omega) (cos omega)       ]]]
  [assume q (sample (gamma-dist 10. 100.))]
  [assume Q (* (eye 2) q)]
  [assume x 
    (mem (lambda (t)
      (if (< t 1)
        [1. 0.]
        (sample (mvn-dist (mmul A (x (dec t))) W)))))]
  [observe (mvn (mmul C (x 1)) R) [...]] 
  ...
  [observe (mvn (mmul C (x 100)) R) [...]] 
  [predict omega]
  [predict q]
\end{verbatim}}
Here {\tt [...]} refers to a vector or matrix literal.
% ...

We compare results for PGAS with 10 particles to ICSMC with $10,20,50,100,200,$ and $500$ particles.
In each case we run 100 PMCMC sweeps and 25 restarts with different random seeds.
To characterize mixing rates we calculate the effective sample size (ESS) of the aggregate sample set $\{w_t^{s,l},\vec{z}_t^{s,l}\}$ over all sweeps $s = \{1, \ldots, 100\}$,
\begin{align*}
    {\rm ESS}_t
    &=
    \frac{1}{\sum_k (V_t^k)^2}
    ,
    &
    V_t^k 
    &= 
    \sum_{s=1}^{100} \sum_{l=1}^L w^{s,l}_t I[\vec{z}_t^k = \vec{z}_t^{s,l}]
    .
\end{align*}
Here $V_t^k$ represents the total importance weight associated with each unique value $\vec{z}_t^k$ in $\{\vec{z}_t^{s,l}\}$.

Figure \ref{fig:lgm-ess} shows the ESS as a function of $t$. 
ICSMC shows a decreasing ESS as $t$ approaches 0, indicating poor mixing for values sampled at early generations. 
PGAS, in contrast, exhibits an ESS that fluctuates but is otherwise independent of $t$.
ESS estimates varied approximately 15\% relative to the mean across independent restarts.
This suggests that fluctuations in the ESS reflect variations in the prior probability of latent transitions $\p{\vec{z}_t}{\vec{z}_{t-1}, \vec{A}}$.

For this model, our PGAS implementation with 10 particles has a computational cost per sweep comparable to that of ICSMC with 300 particles. 
However, when we consider the ESS per computation time, the increases in mixing efficiency outweigh increases in computational cost for state estimates below $t \simeq 50$.    
This is further illustrated in Figure \ref{fig:lgm-omega-q}, which shows the standard deviation of sample estimates of the parameters $\omega$ and $q$.
PGAS shows better convergence per sweep, particularly for estimates of $q$.
ICSMC with 500 particles performs similarly to PGAS with 10 particles when estimating $\omega$, though ICSMC with 500 particles notably has a higher cost per sweep.

\begin{figure}[!t]
\includegraphics[width=3.3in]{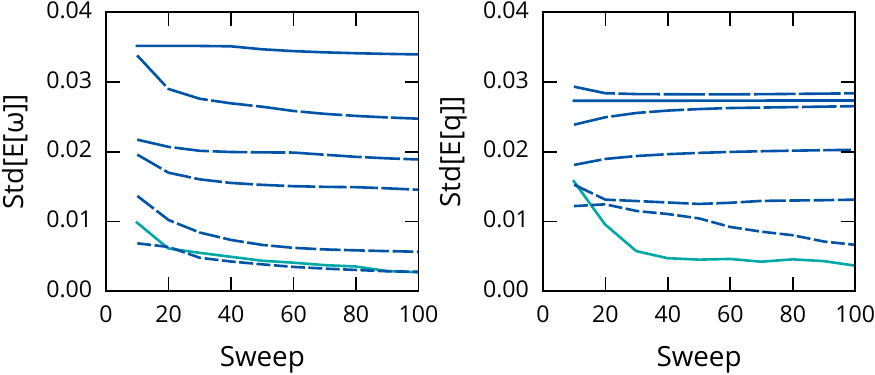}
\caption{\label{fig:lgm-omega-q} Standard deviation of posterior estimates $E[\omega]$ and $E[q]$ over 25 independent restarts, as a function of the PMCMC sweep. PGAS results are shown in cyan, ICSMC results are shown in blue, with shorter dashes indicating a larger particle count.}
\end{figure}

\section{Discussion}



Relative to other PMCMC methods such as ICSMC, PGAS methods have qualitatively different mixing characteristics, particularly for variables sampled early in a program execution.
Implementing PGAS in the context of probabilistic programs poses technical challenges when programs can make use of recursion and memoization.
The technique for traced evaluation developed here incurs an additional computational overhead, but avoids unnecessary recomputation during regeneration.
When the cost of recomputation is large this will result in computational gains relative to a naive implementation that re-executes the suffix fully. 
Note that our approach tracks upstream, not downstream dependencies.
In other words, we know what environment variables affect the value of a given expression, but not which expressions in a suffix depend on a given variable.
All referenced symbol values must therefore be checked during regeneration, which can require a $O(LN^2)$ computation in itself.
Further gains could be obtained constructing a downstream dependency graph for the suffix, allowing more targeted regeneration via graph walk techniques analogous to those employed in Venture \cite[]{mansinghka_arxiv_2014}.
At the same time, the empirical results presented here are indicative of the fact that, even without these additional optimizations, PGAS can easily yield better statistical results in cases where the parameter space is large and ICSMC sampling fails to mix.

\section{Acknowledgements}

We would like to thank our anonymous reviewers, as well as Brooks Paige and Dan Roy for their comments on this paper. 
JWM was supported by Google and Xerox. 
HY was supported by the EPSRC. 
FW and VKM were supported under DARPA PPAML. 
VKM was additionally supported by the ARL and ONR.

\bibliographystyle{plainnat}
\bibliography{pgas}

\end{document}